\documentclass[conference]{IEEEtran}
\IEEEoverridecommandlockouts
\usepackage{cite}
\usepackage{amsmath,amssymb,amsfonts}
\usepackage{algorithmic}
\usepackage{graphicx}
\usepackage{textcomp}
\usepackage{xcolor}
\def\BibTeX{{\rm B\kern-.05em{\sc i\kern-.025em b}\kern-.08em
    T\kern-.1667em\lower.7ex\hbox{E}\kern-.125emX}}
\begin{document}

\title{Graph-Eq: Discovering Mathematical Equations using Graph Generative Models }

\author{Nisal Ranasinghe\textsuperscript{\rm 1}, Damith Senanayake\textsuperscript{\rm 1}, Saman Halgamuge\textsuperscript{\rm 1} \\
\textsuperscript{\rm 1}AI, Optimization and Pattern Recognition Research Group, Dept. of Mechanical Eng., \\ University of Melbourne, Australia}

\maketitle

\begin{abstract}
The ability to discover meaningful, accurate, and concise mathematical equations that describe datasets is valuable across various domains. Equations offer explicit relationships between variables, enabling deeper insights into underlying data patterns. Most existing equation discovery methods rely on genetic programming, which iteratively searches the equation space but is often slow and prone to overfitting. By representing equations as directed acyclic graphs, we leverage the use of graph neural networks to learn the underlying semantics of equations, and generate new, previously unseen equations. Although graph generative models have been shown to be successful in discovering new types of graphs in many fields, there application in discovering equations remains largely unexplored. In this work, we propose Graph-EQ, a deep graph generative model designed for efficient equation discovery. Graph-EQ uses a conditional variational autoencoder (CVAE) to learn a rich latent representation of the equation space by training it on a large corpus of equations in an unsupervised manner. Instead of directly searching the equation space, we employ Bayesian optimization to efficiently explore this learned latent space. We show that the encoder-decoder architecture of Graph-Eq is able to accurately reconstruct input equations. Moreover, we show that the learned latent representation can be sampled and decoded into valid equations, including new and previously unseen equations in the training data. Finally, we assess Graph-Eq's ability to discover equations that best fit a dataset by exploring the latent space using Bayesian optimization. Latent space exploration is done on 20 dataset with known ground-truth equations, and Graph-Eq is shown to successfully discover the grountruth equation in the majority of datasets.  
\end{abstract}


\section{Introduction}

Mathematical equations play a crucial role in our understanding of real world phenomena across many scientific and engineering domains. By expressing relationships between variables in an interpretable form, equations provide insights into the structure of data. However, discovering of equations that describe these phenomena is not always straightforward. Traditionally, equations are derived using domain expertise by building upon theoretical principles and empirical observations. However, this requires theoretical expertise and intuition, and in many cases require the use of simplifying assumptions due to mathematical complexity. This may result in oversimplifying these equations, and overlooking of complex hidden pattern within the data. 

The rapid growth in data availability across many domains has opened new opportunities for data-driven equation discovery. With access to vast amounts of high-resolution data, researchers have the opportunity to discover accurate mathematical equations without relying on domain expertise. Advancements in machine learning has accelerated this, with many work done in creating efficient algorithms that can find equations that accurately describe data \cite{brunton_discovering_2015, sahoo_learning_2018, martius_extrapolation_2016, long_pde-net_2018, kamienny_deep_2022, schmidt_distilling_2009}. This area is sometimes referred to as symbolic regression (SR), and involves searching through the space of equations until a well fitting equation is found. More formally, the objective of symbolic regression is to discover an equation that can map the features $\textbf{X} \in \mathbb{R}^d$ to the output $y \in \mathbb{R}$ using a dataset of  $\textbf{X}$ and $y$ pairs. 

Most early symbolic regression methods involved the use of genetic programming \cite{koza_genetic_1994, dubcakova_eureqa_2011, schmidt_distilling_2009}. Genetic programming explores the space of possible equations by evolving mathematical expressions using operators such as mutation, crossover, and selection. While this approach has been successful in discovering interpretable equations, it often suffers from inefficiencies. Genetic programming can be computationally expensive, particularly when dealing with large datasets or complex equation spaces. Additionally, these methods are prone to overfitting, producing overly complex equations that fail to generalize well to unseen data. 

With recent advancements in deep learning, some deep learning-based symbolic regression methods have emerged \cite{kamienny_deep_2022, valipour_symbolicgpt_2021, biggio_neural_2021, kusner_grammar_2017}. These methods leverage the powerful representation learning capabilities of neural networks to explore the space of possible equations. Unlike genetic programming, deep learning models can learn rich feature representations from data, enabling nonlinear relationships effectively while generalizing well to unseen data. Techniques such as recurrent neural networks (RNNs), transformer architectures, and variational autoencoders (VAEs) have been adapted for symbolic regression tasks. Models like SymbolicGPT \cite{valipour_symbolicgpt_2021} and NeSymRes \cite{biggio_neural_2021} utilize transformer-based architectures to generate symbolic expressions, while other approaches like Deep Symbolic Regression \cite{petersen_deep_2021} combine neural networks with reinforcement learning to improve search efficiency. Deep learning-based methods have shown promising results in discovering interpretable equations from data, making them increasingly popular in SR.

Generative models such as variational autoencoders \cite{kingma_auto-encoding_2022}, Generative adversarial networks \cite{goodfellow_generative_2014}  and transformer-based models  \cite{vaswani_attention_2017} have been shown to be effective in learning powerful representations of data using unsupervised training pipelines. Although most of these generative models were originally proposed for image generation, they have recently been adapted for graph generation tasks \cite{cao_molgan_2022, zhang_d-vae_2019, dong_pace_2022}. These methods can not only learn powerful representations of graph datasets, but can also enforce some regularity in the latent representation, allowing latent space exploration to discover new, previously unseen graphs. 

Mathemetical equations can be effectively represented using graphs, in particular directed acyclic graphs (DAG). This enables the use of powerful graph representation learning methods to learn representations of equations, and perform equation discovery using these latent representations. However, the use of graph generative models for equation discovery remains largely unexplored.

In this work, we propose Graph-EQ, a graph generative model designed for efficient and accurate equation discovery. Unlike traditional symbolic regression methods that search directly through the space of equations, Graph-EQ leverages a conditional variational autoencoder (VAE) to learn a structured latent representation of the equation space. By mapping equations into a continuous latent space, Graph-EQ enables efficient exploration using Bayesian optimization, improving search efficiency. This approach allows Graph-EQ to capture mathematical relationships while reducing the risk of overfitting. To the best of our knowledge, this is the first method that uses graph neural networks for equation discovery. 

\begin{figure*}[t]
    \centering
    \includegraphics[width=1\textwidth]{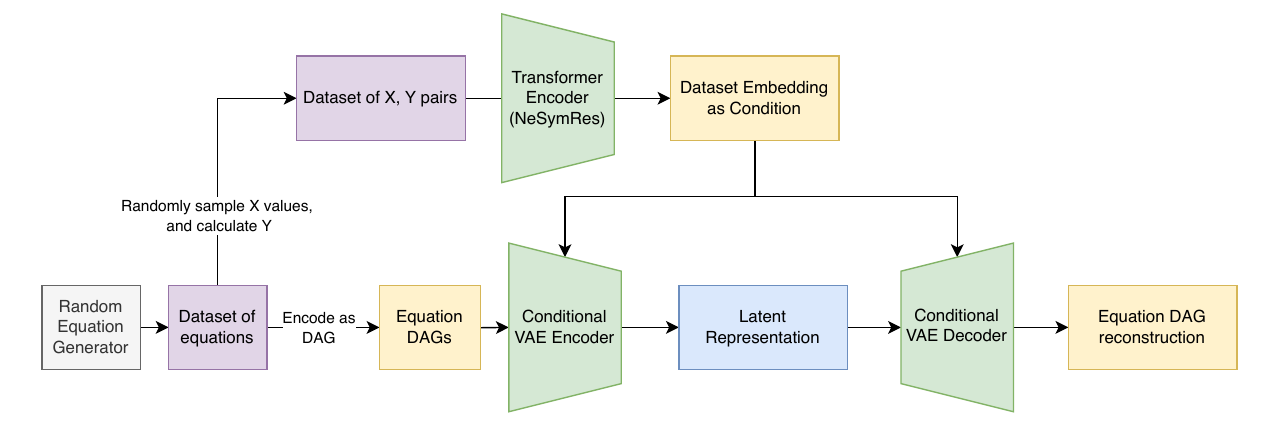}
    \caption{The architecture of Graph-Eq. The random equation generator generates equations for training. These equations are represented as equations DAGs and parallely used to create a small dataset of $\textbf{x}, y$ pairs. This dataset is then used to calculate a dataset embedding to be used to condition the VAE encoder and decoder. The VAE decode }
    \label{fig:graph-eq}
\end{figure*}

The main contributions of this paper are,
\begin{itemize}
    \item A graph-based encoding method that represents mathematical equations as graph structures, enabling the use of graph neural networks to capture the semantics of equations
    \item Graph-Eq: A conditional variational autoencoder (VAE) that can learn a continous latent representation of equations, enabling the search of equations in a continous space.
    \item An evaluation of the quality of Graph-EQ's latent space representation, and its effectiveness in discovering best-fitting equations using Bayesian optimization.
\end{itemize}

The rest of the paper is organized as follows. In Section \ref{sec:related_work} we review related literature related to data driven discovery of equations. Then, in Section \ref{sec:method} we introduce Graph-Eq, detailing our graph based equation encoding, the VAE based architecture for learning latent representations of equations and the Bayesian optimization based search strategy for equation discovery. Next, we present the results of our experiments in Section \ref{sec:results} and summarize our finding in Section \ref{sec:conlusion}.

\section{Related Work}
\label{sec:related_work}

\textbf{Symbolic regression} (SR) refers to a class of algorithms that aim to discover mathematical expressions that best describe a dataset. Unlike traditional regression which fits data to a pre-defined model, SR methods search the space of possible mathematical equations to identify the most suitable one. SR is particularly challenging since the number of possible equations grows exponentially with the complexity of expressions. Searching through all possible equations has been proven to be an NP-Hard problem \cite{virgolin_symbolic_2022}. Therefore, SR methods focus on efficiently exploring the space of equations. Though SR was originally proposed as a genetic programming problem \cite{koza_genetic_1994}, advancements in machine learning has led to a wide variety of methods being proposed for SR. This includes many neural network based methods in the recent past \cite{ranasinghe_ginn-lp_2024, ranasinghe_ginn-kan_2024, biggio_neural_2021, valipour_symbolicgpt_2021}.

\textbf{Variational Autoencoders} (VAE) are a type of generative model that learn a probabilistic mapping between data points and a lower-dimensional latent space, and a probabilistic model that can decode points from the latent space into new data points \cite{kingma_auto-encoding_2022}. This is done using an encoder decoder architecture. The encoder learns an approximate probabilistic mapping $q_{\phi}(z | x)$ where $x$ is the input data vector, $z$ is the latent vector and $\phi$ are the parameters of the encoder. The decoder learns a probabilistic generative model $p_{\theta}(x | z)$ where $\theta$ are the parameters of the decoder. VAE's regularize the latent space by introducing a prior distribution, typically a multivariate Gaussian, which encourages the learned latent representations to follow a structured distribution. This helps promote smoothness of the latent space, resulting in the latent representations following a probability distribution. Therefore, it is possible to sample points from the latent space and decode them into meaningful data points that resemble the original data distribution.

\textbf{Conditional VAE} (CVAE) is an extension of the standard VAE designed to generate data conditioned on specific input variables \cite{sohn_learning_2015}. CVAE's allow the introduction additional conditioning information such as labels, features or extra context, to guide both the encoding and decoding processes. The encoder in a CVAE's learns a conditional probabilistic mapping $q_{\phi}(z | x, c)$ where $c$  is the condition vector. Similarly, the decoder learns a conditional probabilistic generative model $p_{\theta}(x | z, c)$. 

\textbf{Bayesian optimization} (BO) is an optimization technique for optimizing black-box functions that are expensive to evaluate. It is particularly well-suited for scenarios where the objective function is costly, non-convex, or lacks an explicit mathematical form. Bayesian optimization builds a probabilistic model, often a Gaussian process (GP), to approximate the unknown objective function. This surrogate model is iteratively refined as new data points are sampled, enabling the algorithm to focus on promising regions of the search space efficiently. BO is ideal for exploring structured latent spaces, to find data points that are that are optimal for a given task. Many works have demostrated the effectiveness of using BO to explore latent spaces learned by generative models such as VAEs. 

\textbf{Graph neural networks} (GNN) are a class of neural networks designed to operate on graph data. They have been shown to be highly effective in learning the relational information within graphs and performing tasks such as node classification, link prediction and graph regression. GNNs use a mechanism called message passing, which allows nodes to exchange information with their neighbouring nodes, iteratively updating each node's representation. This allows GNNs to learn representations of graphs in a permutation invariant manner, ensuring that the order of nodes does not affect the learnings of the network. 

\textbf{Deep graph generative models} expand on the capabilities of GNNs by not only learning representations of graphs, but also allowing the generation of new, previously unseen graphs that resemble the original data distribution of graphs. The latent representations learnt by these models can then be explored to discover new graphs that are optimal for a given task. Combined with black box optimization techniques such as BO, graph generative models have been shown to be successful in discovering optimal neural architectures \cite{zhang_d-vae_2019}, new molecular structures \cite{cao_molgan_2022} and proteins \cite{guo_generating_2021}.

\section{Method}
\label{sec:method}

In this work, we introduce Graph-Eq, a conditional VAE-based deep graph generative model for learning representations of mathematical equations. Graph-Eq is trained on a large number of randomly generated equations in a completely unsupervised manner to learn a rich latent representation. Once trained, the latent space learnt by Graph-Eq can be explored using black-box optimization techniques to discover equations that best fit tabular datasets. The architecture of Graph-Eq is illustrated in Figure \ref{fig:graph-eq}.

We define the task of equation discovery as follows. Let $\mathcal{D}$ be a dataset consisting of input-output pairs of $\textbf{x}$ and $y$ where $\textbf{x} \in \mathbb{R}^d$ represents the input features and $y \in \mathbb{R}$ is the corresponding output. The goal is to identify a mathematical equation $f : \mathbb{R}^d \rightarrow \mathbb{R}$ such that $y \approx f(\textbf{x})$ for all points in the dataset. The desired equation should be both \textit{accurate}, minimizing prediction error on the data, and \textit{interpretable}, ideally expressed in a concise symbolic form.  

To achieve this, Graph-Eq operates in two stages
\begin{enumerate}
    \item Training stage - Unsupervised training of the Graph-Eq on a large equation dataset, to learn a rich latent representation of equations
    \item Exploration stage - For a given dataset of $\textbf{x}, y$ pairs, the latent space is explored using Bayesian optimization, till we find an equation providing the least possible root mean squared error.
\end{enumerate}

\subsection{Training dataset}

As with many deep generative models, Graph-Eq needs to be trained on a large equation dataset to learn a well structured latent space. Since such a large dataset cannot be feasibly obtained from a real-world equation dataset, we instead use a data generator to sample random equations for training. We use the framework introduced by (Lample \& Charton, 2020) \cite{lample_deep_2020} to generate random equations. 

Once the equations are generated, they need to be converted into DAG structures so that they can be passed into the graph neural network. We represent equations as DAGs using the following graph representation.

\begin{itemize}
    \item Input features $x_i$ as source nodes
    \item Operands as intermediate nodes
    \item Output $y$ as the sink node
\end{itemize}

\begin{figure}[t]
    \centering
    \includegraphics[width=1\columnwidth]{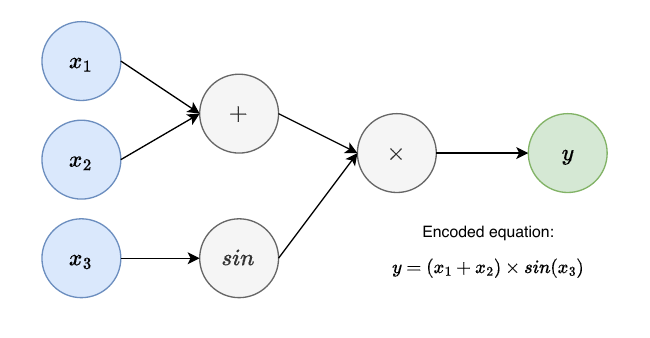}
    \caption{An example DAG representation of an equation. The intermediate nodes represent operators while the source and sink nodes represent inputs and outputs respectively.}
    \label{fig:equation-dag}
\end{figure}

This representation encodes the mathematical operations of an equation within the DAG structure. An example equation encoded as a DAG is shown in Figure \ref{fig:equation-dag}. 

\subsection{Graph Conditional VAE}

We then use a graph-based conditional VAE to learn a structured latent representation of these equation DAGs, which will later allow us to sample the latent space and generate new, previously unseen equation DAGs. Graph-Eq follows the D-VAE architecture \cite{zhang_d-vae_2019}, which is a GNN based VAE for encoding computations of DAGs into a latent representation. D-VAE uses an asynchronous message passing algorithm, which aids in encoding not just the structure of the DAGs but also the computational flow. This message passing technique respects the order of nodes, enabling it to learn the directional information within the DAG. This is particularly well suited for encoding equations, since an equation would represent a computation on the equation DAG.

However, simply using D-VAE to learn equation representations poses a key challenge. To efficiently perform latent space optimization for discovering optimal equations, the latent space needs to be organized not only according to the structural similarity of equations, but also according to the functional similarity of the equations. Structural similarity ensures that equations with similar computational graphs are mapped closely in the latent space, while functional similarity ensures that equations producing similar outputs under various inputs are also grouped together. Without this, equations with vastly different behaviors might be encoded similarly, making optimization ineffective. However, D-VAE does not allow incorporating this information into the training process, since it optimizes the reconstruction loss and VAE regularization loss, which primarily captures the structural information of the input graphs.

To address this challenge, we propose the use of a conditional VAE, in place of a traditional VAE architecture. The VAE component of Graph-Eq conditions the latent space to ensure that functionally similar equations are mapped close together, in addition to structurally similar ones. This promotes a more meaningful organization of the latent space, where equations that yield similar outputs under various inputs are positioned near each other. By incorporating functional similarity into the encoding process, we hope to facilitate more efficient discovery of optimal equations. 

\subsection{Encoding functional aspects of equations using dataset embeddings}

We note that functional aspects of equations are inherently encoded in any dataset that is generated using the equation, as the data points reflect the underlying functional relationship between variables. We use this insight when designing the input condition to the conditional VAE. For each equation, we create a small tabular dataset by randomly sampling 500 $\textbf{x}$ values and calculating corresponding output $y$ values. We then represent each of these datasets using a low dimensional dataset embedding. 

As a baseline embedding, we fit a polynomial of fixed degree to the dataset of input-output pairs and use the polynomial coefficients as the dataset embedding. This method provides a simple yet effective way to capture basic functional characteristics of the dataset. For instance, the polynomial coefficients encode trends such as linearity, curvature, and higher-order behaviors, offering a compact representation of the equation’s structure. While this method may struggle with complex functional forms like trigonometric or exponential patterns, it serves as a useful comparison point for more sophisticated embeddings.

In addition to the polynomial embedding, we also leverage a pre-trained Set Transformer \cite{lee_set_2019} encoder to calculate a more descriptive embedding for the dataset. We use the pre-trained Set Transformer encoder from NeSymRes \cite{biggio_neural_2021}, which is trained on datasets generated using 100 million equations. The set transformer architecture is particularly suited for learning latent representations of data, since they are able to effectively encode unordered sets of data points. Since sampled data is inherently unordered, the permutation invariant nature of the Set Transformer ensures that the embedding does not depend on which order the data is sampled.

The resultant dataset embedding captures the functional properties of the underlying equation, since the dataset is generated directly from the equation. Therefore, we expect the the embedding would encode key characteristics of the equation such as linearity, periodicity or polynomial degree. By conditioning the VAE on this dataset embedding, we guide the encoder to organize equations in a way that reflects both their structural form and functional behaviour.

\subsection{Decoding the latent space into equation DAGs}

Once the conditional VAE encodes the equation DAG onto a point in the latent space, the next step is to decode them into equation DAGs. The conditional VAE decoder learns a probabilistic generative model that maps the points in the latent space, conditioned by the dataset embedding into an equation DAG. This is done by first mapping the latent vector to a hidden state using a multi-layer perceptron (MLP). This hidden state is then fed to a GRU layer, which generates the reconstructed DAG node by node. The GRU predicts the type distribution of each node, and samples them to create each node of the DAG. This is continued until an ending node is generated, or until a maximum number of nodes is reached. 

\subsection{Training Graph-Eq}

During the training stage, we train Graph-Eq to accurately reconstruct the input equation DAGs at the output. Similar to D-VAE \cite{zhang_d-vae_2019}, we use teacher forcing to measure the reconstruction loss. The loss is calculated as,

\begin{equation}
    loss = reconstruction\_loss + \alpha * KL\_Divergence
\end{equation}

Here, $\alpha = 0.005$. The $KL\_Divergence$ term regularizes the latent space by ensuring that the learned distribution remains close to the prior distribution, preventing overfitting and encouraging smooth interpolation between points in the latent space. We use $\mathcal{N}(0, I)$ as the prior distribution, which is commonly used when training VAEs. We train the model to minimize the loss function using mini-batch SGD with Adam optimizer. Across all experiments, Graph-Eq is trained for 100 epochs, with a batch size of 32.

All experiments were conducted on a high performance computing cluster where each experiment used a 32-core, 2.90GHz Intel Xeon Gold 6326 CPU and a single NVIDIA A-100 GPU.

\subsection{Exploring latent space for equation discovery}

\begin{figure*}[t]
    \centering
    \includegraphics[width=1\textwidth]{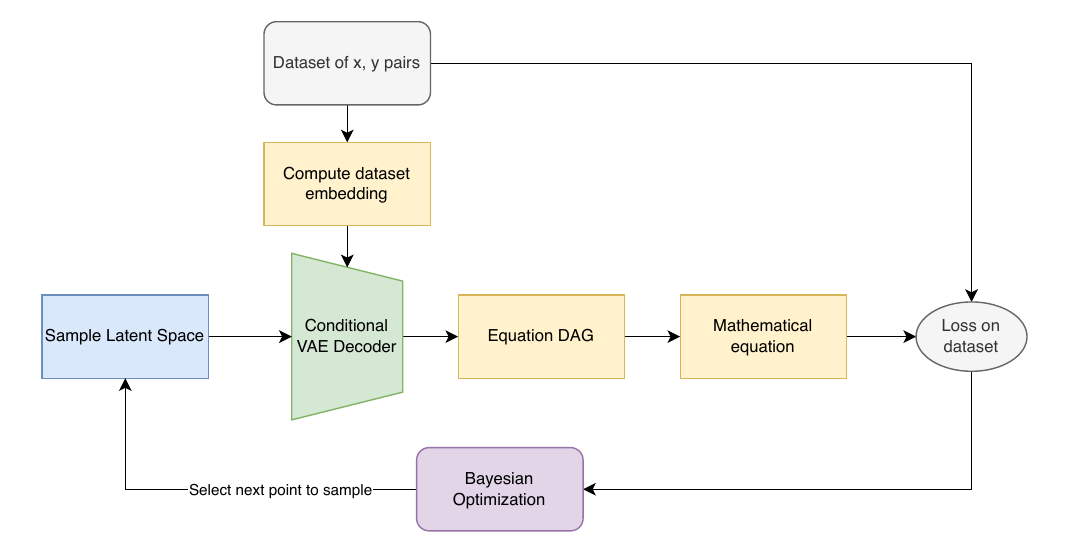}
    \caption{The equation discovery pipeline of Graph-Eq. Bayesian optimization is used to efficiently sample points in the latent space until an optimal equation is discovered.}
    \label{fig:inference}
\end{figure*}

Once Graph-Eq is trained and a good latent representation of equations is learnt, the latent space can be sampled to generate new equations that resemble the original distribution of equations. This also includes new equations that were previously unseen by the model. This latent space, along with the Graph-Eq decoder can be used to discover optimal equations that describe datasets. Given a dataset of input-output $\textbf{x}, y$ pairs, the objective is to identify a point in the latent space which corresponds to an equation that best fits the dataset. We do this by exploring the latent space and decoding the latent space vectors in the equation DAGs. The exploration continues until a satisfactory equation is discovered, which well fits the data. Although exploring the latent space can be done naively, by simply searching points on a fine grid, this appraoch is inefficient and computationally expensive. Bayesian optimization is particularly well suited towards these types of black-box optimization problems. It can balance exploration and exploitation by building a probabilistic model of the objective function, and using that to strategically choose which point to evaluate next while also updating the probabilistic model. This allows the search process to focus on promising regions of the latent space, significantly reducing the number of candidate equations that need to be evaluated. We use Bayesian optimization on the learnt latent space to discover best fitting equations for a given dataset, as illustrated in Figure \ref{fig:inference}.

To guide Bayesian optimization during the latent space exploration, we optimize the following $score$ metric.
\begin{equation}
    score = \frac{1}{1 + MSE}
    \label{eq:error}
\end{equation}

where MSE is the mean squared error between the values predicted by the equation decoded using the sampled latent space vector, and actual output values in the dataset. This metric is chosen because it maps the MSE to a bounded range between 0 and 1, also ensuring that lower MSE values result in higher scores. By transforming the MSE in this way, we provide Bayesian optimization with a smoother objective function that better distinguishes between candidate solutions with small performance differences. Additionally, this formulation encourages the optimizer to focus on minimizing MSE while maintaining numerical stability, preventing extreme values from dominating the search process.

\section{Results and Discussion}
\label{sec:results}

We evaluate the performance of Graph-Eq through two key assessments: (1) examining the quality of the learnt latent space, and (2) evaluating its ability to recover equations that describe given datasets.

We generate two sets of equations, consisting of 20,000 and 120,000 randomly generated equations. These datasets are used to train Graph-Eq in an unsupervised manner so that a good latent representation is learned. We train separately on each of the datasets, to evaluate how the size of the dataset affects training performance. The generated equations consist of equations that use a diverse set of mathematical operators: 1. Addition 2. Multiplication 3. Subtraction 4. Division 5. Square root 6. $\log$ 7. Exponential 8. $\sin$ 9. $\cos$ 10. $\tan$ 11. $\arcsin$ 12. Power. After generating the equation dataset using the data generator, we hold out 10\% of it as the test data.

For each dataset, we perform the following experiments
\begin{enumerate}
    \item Graph-Eq with original D-VAE (no conditional VAE)
    \item Graph-Eq with Conditional VAE, conditioned on the dataset embedding
\end{enumerate}

\begin{table*}[t]
    \centering
    \caption{The results of Graph-Eq after unsupervised training on randomly generated equations. We calculate the reconstruction accuracy on the test set. The validity, uniqueness and novelty measure the quality of the latent space, and are calculated after randomly sampling 1,000 points from the latent space, and decoding them into equation DAGs}
    \begin{tabular}{|l|l|l|l|l|l|}
    \hline
        \textbf{Dataset Size} & \textbf{Dataset Embedding} & \textbf{Reconstruction accuracy} & \textbf{Validity} & \textbf{Uniqueness} & \textbf{Novelty}  \\ \hline
        20K & None (Vanilla D-VAE) & 74.48 & 66.51 & 28.94 & 73.27  \\ \hline
        20K & Polynomial fitting-based embedding & 74.43 & 76.62 & 29.5 & 68.12 \\ \hline
        20K & NeSymRes - Mean aggregation & 74.77 & 78.32 & 33.5 & 72.61  \\ \hline
        20K & NeSymRes – MLP projection (5120 $\rightarrow$ 5) & 72.68 & 67.93 & 39.84 & 74.11  \\ \hline 
        20K & NeSymRes – MLP projection (5120 $\rightarrow$ 10) & 60.05 & 81.33 & 35.31 & 67.47  \\ \hline \hline
        120K & None (Vanilla D-VAE) & 84.15 & 84.81 & 38.2 & 51.89  \\ \hline
        120K & Polynomial fitting-based embedding & 83.84 & 72.61  & 38.47 & \textbf{54.88} \\ \hline
        120K & NeSymRes – 10 & 84.6 & \textbf{87.36} & 36 & 48.45  \\ \hline
        120K & NeSymRes - MLP projection (5120 $\rightarrow$ 5) & 84.96 & 80.49 & 37.09 & 50.57 \\ \hline
        120K & NeSymRes - MLP projection (5120 $\rightarrow$ 10)   & \textbf{85.09} & 80.89 & \textbf{40.13} & 54.65  \\ \hline
    \end{tabular}
    \label{tab:train_results}
\end{table*}

\subsection{Visualizing the latent space}

\begin{figure}[t]
    \centering
    \includegraphics[width=1\columnwidth]{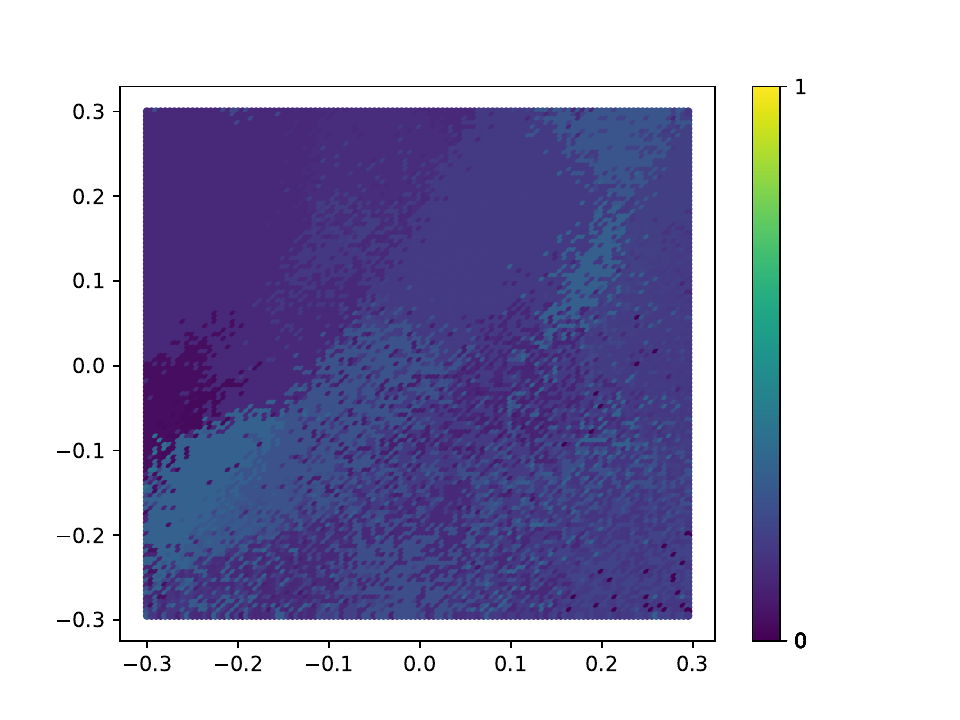}
    \caption{A 2D subspace of the latent space learnt by Graph-Eq is visualized to demonstrate the smoothness of the latent space for a single SR dataset. Each point in this subspace is decoded into an equation DAG, and the $1/(1 + MSE)$ score is visualized in the colourmap.}
    \label{fig:latent_space}
\end{figure}

As outlined in Section \ref{sec:method}, we optimize the score metric defined in Equation \ref{eq:error} during Bayesian Optimization (BO) for equation discovery. For BO to effectively explore the latent space, the space should ideally be smooth with respect to the error metric — meaning that similar points in the latent space should correspond to equations with similar error values.

To illustrate this smoothness, we visualize the score landscape for a single dataset in a reduced-dimensional latent space. Specifically, we project the latent space onto a 2D subspace spanned by the first two principal components. We then uniformly sample points on a grid within this 2D subspace, decode them into equation DAGs, and use these equations to predict target $y$ values. The corresponding error values are then computed and visualized using a color map over the 2D latent space. 

We observe that the latent space exhibits some smoothness with respect to the score metric, as shown in Figure \ref{fig:latent_space}. This smoothness is desirable for BO, since it enables the optimizer to identify promising high score regions of the latent space by levaraging the continuity of the space. This highlights Graph-Eq's ability to not only capture meaningful information from the equation DAGs, but also organize it's representation to better suit the task of equation discovery. 

\subsection{Training results}

After training Graph-Eq, we evaluate the quality of the VAE encoder and decoder using the reconstruction accuracy. The proportion of correctly reconstructed equation DAGs is measured by comparing the reconstructed DAG structure with the original input DAG. A reconstruction is considered correct if the predicted DAG is structurally identical to the original, including the correct placement of operators and input variables. This is evaluated on the test dataset. 

Moreover, following D-VAE \cite{zhang_d-vae_2019} we assess the quality of the learnt latent space using the following metrics. These metrics are calculated after randomly sampling 1,000 points from the latent space. 
\begin{itemize}
    \item Percentage of valid equations (Validity) - Measures the proportion of decoded equations that are syntactically correct and mathematically valid. For example, if a decoded equation DAG has a $\sin$ operator with two input connections, this is mathamatically invalid. 
    \item Percentage of unique equations (Uniqueness) - Measures the diversity of decoded equations, by calculating the percentage of distinct equations discovered by randomly smapling the latent space
    \item Percentage of equations unseen in training data (Novelty) - Evaluates the model's ability to generalize, by calculating the number of novel equations discovered which were not part of the training data.
\end{itemize}

\begin{table*}[t]
    \centering
    \caption{Equation discovery results}
    \label{tab:eq_discovery}
    \begin{tabular}{|l|l|l|l|}
    \hline
        \textbf{Training dataset size} & \textbf{Train epochs} & \textbf{Dataset Embedding} & \textbf{Solution rate}  \\ \hline
        120K  & 100 & None (Vanilla D-VAE) & 40\% \\ \hline
        120K & 100 & NeSymRes - MLP projection (5120 $\rightarrow$ 5)  & \textbf{55\%} \\ \hline
        120K & 100 & NeSymRes - MLP projection (5120 $\rightarrow$ 10)  & 50\% \\ \hline
    \end{tabular}
\end{table*}

We train Graph-Eq on datasets containing 20K and 120K equations and present the results in Table \ref{tab:train_results}. The dataset embeddings are generated using the pre-trained Set Transformer encoder from NeSymRes \cite{biggio_neural_2021}, which produces a $512 \times 10$ dimensional 2D embedding. Since the conditional VAE requires a one-dimensional condition vector as input, we apply dimensionality reduction to the dataset embeddings. 

The embeddings we use are as follows.

\begin{itemize}
\item Polynomial fitting-based embedding: Fitting a polynomial of fixed degree to the dataset of input-output pairs and using the polynomial coefficients as the dataset embedding.
\item NeSymRes - Mean aggregation: Averaging across the first dimension of the embedding to produce a 10-dimensional vector.
\item NeSymResMLP projection (5120 $\rightarrow$ 5): Flattening the embedding into a 5120-dimensional vector, followed by an MLP to reduce it to 5 dimensions.
\item NeSymRes - MLP projection (5120 $\rightarrow$ 10): Flattening the embedding into a 5120-dimensional vector, followed by a MLP to reduce it to 10 dimensions.
\end{itemize}

As expected, the larger training dataset size results in a better performance across all experiments. This is because deep learning methods like Graph-Eq can learn more meaningful latent representations when trained with larger datasets. 

For the larger dataset, the conditional VAE with dataset embeddings performs better across all performance metrics, demonstrating the effectiveness of incorporating the dataset embedding as a condition when training the VAE. 

The polynomial fitting-based embedding achieves the highest novelty performing marginally better than the next best method. It's likely that the more descriptive embedding of NeSymRes results in the other methods slightly overfitting to the training data. This would result in lower novelty, since the latent space would better capture the semantics of the equations within the training data. However, the NeSymRes embeddings with perform better across reconstruction accuracy, validity and uniqueness, indicating that the dataset embedding improves the performance of the VAE, as well as the quality of the latent space. A higher validity score shows that these versions of Graph-Eq are successful in learning the structural semantics of equations, including structural constraints. A higher uniqueness indicates that the latent space captures a more diverse set of equations. Moreover, we observer that the MLP based dimensionality reduction of the dataset embedding increases most of the performance metrics, compared against the simple aggregation-based dimensionality reduction. This could be due to some of the information being lost when using a simple aggregation method to reduce the dimensionality of the embedding. We also note that reducing the dimensionality to 5 performs slightly worse than reducing it to 10. This can be attributed to the fact that higher dimensional embeddings can retain more information when compared to lower dimensional embeddings. However, both the MLP based dimensionality reduction methods perform worse than the simple aggregation method in terms of the validity. This means that VAE with the MLP based dimensionality reduced embedding is a bit worse in learning the structural semantics of equations, leading to the generation of more invalid equations. 

\subsection{Equation discovery results}

To evaluate Graph-Eq’s performance in equation discovery, we generate 20 additional unseen equations not present in the training data. For each of these equations, we create a dataset containing 10,000 randomly sampled input-output pairs. The goal is to explore the latent space using Bayesian Optimization (BO) and identify the equation that best fits the dataset, ideally recovering the original equation that generated the data.

For each dataset, we perform 10 iterations of BO across 10 trials, and obtain the best result across all 10 trials. When performing BO, we minimize the error defined in Equation \ref{eq:error}. The equation that provides the lowest error is selected as the discovered equation. The discovered equation is then simplified using SymPy \cite{meurer_sympy_2017} and then compared against the actual groundtruth equation to determine whether the groundtruth equation has been recovered. After running BO for all 20 datasets, we calculate the number of equations that exactly matched the groundtruth equation. We note that this is a commonly used metric in the SR literature, and is sometimes referred to as the solution rate \cite{la_cava_contemporary_2021}.

The results of the experiments are presented in Table \ref{tab:eq_discovery}. We observe that Graph-Eq with the conditional VAE, conditioned on the NeSymRes based dataset embedding clearly outperforms the vanilla D-VAE, correctly discovering 11 of the groundtruth equations, compared to 8 with the vanilla D-VAE. 

\section{Conclusion}
\label{sec:conlusion}

In this work, we introduced Graph-Eq, a deep graph generative model for equation discovery. By representing mathematical equations as directed acyclic graphs and leveraging the power of graph neural networks and conditional variational autoencoders, Graph-Eq is able to learn a continuous, structured latent representation of mathematical equations. This latent representation can then be explored using black-box optimization techniques such as Bayesian optimization, to discover equations that describe datasets of $\textbf{x}, y$ pairs. 

Our results demonstrate several key advantages of Graph-EQ. First, the incorporation of dataset embeddings as conditions in the VAE training process significantly improves the quality of the learned latent space, as evidenced by higher reconstruction accuracy, validity, and uniqueness metrics compared to when an unconditional VAE is used. Second, Graph-EQ achieves a 55\% solution rate on our test suite, outperforming the baseline model 15\%. This demonstrates the effectiveness of our proposed conditional VAE architecture in capturing both structural and functional similarities between equations.

Despite these promising results, Graph-EQ has notable limitations. One limitation is its inability to discover equations containing numerical constants. For instance, although Graph-Eq can discover the equation $x_1 + x_2$, it cannot discover expressions such as $4.5x_1 + 3x_2$, since the DAG structure we use cannot represent constants. Future work will explore methods to address this limitation.

\section*{Acknowledgement}

This study was supported by the Melbourne Graduate Research Scholarship for the first author, the Australian Research Council grant DP210101135 for the first author and the Australian Research Council grant DP220101035 for the second and the last authors. This research was also supported by The University of Melbourne’s Research Computing Services and the Petascale Campus Initiative. 

\bibliographystyle{IEEEtran}
\bibliography{references}

\end{document}